\documentclass[final]{cvpr}

\usepackage{times}
\usepackage{epsfig}
\usepackage{graphicx}
\usepackage{amsmath}
\usepackage{amssymb}
\usepackage{cite}
\usepackage{bm}
\usepackage{multirow}
\usepackage{booktabs}
\usepackage{arydshln}
\usepackage{pifont}
\usepackage{url}
\usepackage[caption=false]{subfig}
\usepackage{bbm}

\makeatletter
\robustify\@latex@warning@no@line
\makeatother
\usepackage{authblk}

\allowdisplaybreaks[1]

\DeclareMathOperator*{\argmax}{arg\,max}
\DeclareMathOperator*{\argmin}{arg\,min}

\usepackage[pagebackref=true,breaklinks=true,colorlinks,bookmarks=false]{hyperref}

\def\cvprPaperID{} % *** Enter the CVPR Paper ID here
\def\confYear{CVPR 2021}

\begin{document}

%%%%%%%%% TITLE
\title{QPIC: Query-Based Pairwise Human-Object Interaction Detection \\with Image-Wide Contextual Information\vspace{-3ex}}

\author[1]{Masato Tamura}
\author[2]{Hiroki Ohashi}
\author[1]{Tomoaki Yoshinaga}

\affil[1]{Lumada Data Science Lab, Hitachi, Ltd.}
\affil[2]{Center for Technology Innovation - Artificial Intelligence, Hitachi, Ltd.}
\affil[ ]{\textit{\{masato.tamura.sf, hiroki.ohashi.uo, tomoaki.yoshinaga.xc\}@hitachi.com}\vspace{-4ex}}

\maketitle

%%%%%%%%% ABSTRACT
\begin{abstract}
We propose a simple, intuitive yet powerful method for human-object interaction (HOI) detection. 
HOIs are so diverse in spatial distribution in an image that existing CNN-based methods face the following three major drawbacks;
 they cannot leverage image-wide features due to CNN's locality,
 they rely on a manually defined location-of-interest for the feature aggregation, which sometimes does not cover contextually important regions,
 and they cannot help but mix up the features for multiple HOI instances if they are located closely.
To overcome these drawbacks, we propose a transformer-based feature extractor, in which an attention mechanism and query-based detection play key roles. 
The attention mechanism is effective in aggregating contextually important information image-wide, while the queries, which we design in such a way that each query captures at most one human-object pair, can avoid mixing up the features from multiple instances.
This transformer-based feature extractor produces so effective embeddings that the subsequent detection heads may be fairly simple and intuitive.
The extensive analysis reveals that the proposed method successfully extracts contextually important features, and thus outperforms existing methods by large margins (5.37 mAP on HICO-DET, and 5.7 mAP on V-COCO). The source codes
are available at \url{https://github.com/hitachi-rd-cv/qpic}.
\end{abstract}

\vspace{-2ex}

%%%%%%%%% BODY TEXT
\section{Introduction}\label{sec:intro}
\begin{figure}[t]
\centering
\subfloat[]{%
    \label{fig:fig1_heatmap_a}
    \includegraphics[clip,keepaspectratio,width=0.49\columnwidth]{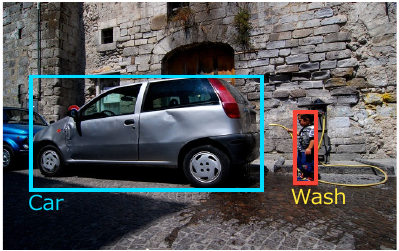}%
}
\subfloat[]{%
    \label{fig:fig1_heatmap_b}
    \includegraphics[clip,keepaspectratio,width=0.49\columnwidth]{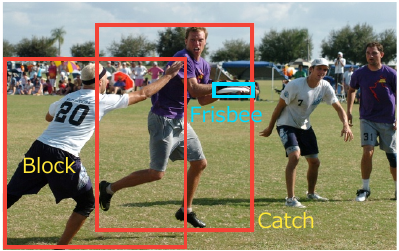}%
} \\
\vspace{-2.0ex}
\subfloat[]{%
    \label{fig:fig1_heatmap_c}
    \includegraphics[clip,keepaspectratio,width=0.49\columnwidth]{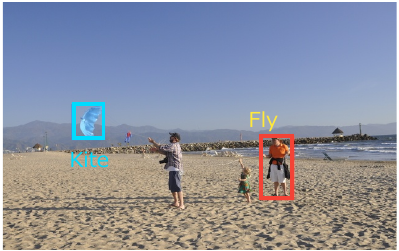}%
}
\subfloat[]{%
    \label{fig:fig1_heatmap_d}
    \includegraphics[clip,keepaspectratio,width=0.49\columnwidth]{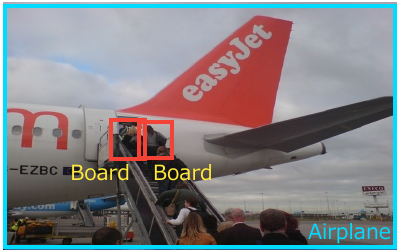}%
}
\caption{Typical failure cases of conventional methods. The ground-truth human bounding boxes, object bounding boxes, object classes, and action classes are drawn with red boxes, blue boxes, blue characters, and yellow characters, respectively.}
\label{fig:fig1_heatmap}
\vspace{-2.0ex}
\end{figure}

Human-object interaction (HOI) detection has attracted enormous interest in recent years for its potential in deeper scene understanding~\cite{chao_wacv2018, gkioxari_cvpr2018, gao_bmvc2018, qi_eccv2018, gupta_iccv2019, li_cvpr2019, wan_iccv2019, wang_iccv2019, zhou_iccv2019, li_cvpr2020, liao_cvpr2020, ulutan_cvpr2020, wang_cvpr2020, zhou_cvpr2020, lin_ijcai2020, yang_ijcai2020, xu_tmm2020, gao_eccv2020, liu_eccv2020, kim_bumsoo_eccv2020, zhi_eccv2020, hai_eccv2020, zhong_eccv2020, kim_dong_eccv2020}. 
Given an image, the task of HOI detection is to localize a human and object, and identify the interactions between them, typically represented as $\langle${\it human bounding box, object bounding box, object class, action class}$\rangle$.

Conventional HOI detection methods can be roughly divided into two types: two-stage methods~\cite{chao_wacv2018, gao_eccv2020, gao_bmvc2018, gkioxari_cvpr2018, gupta_iccv2019, zhi_eccv2020, kim_dong_eccv2020, li_cvpr2020, li_cvpr2019, lin_ijcai2020, liu_eccv2020, ulutan_cvpr2020, wan_iccv2019, hai_eccv2020, xu_tmm2020, yang_ijcai2020, zhong_eccv2020, zhou_iccv2019, qi_eccv2018, wang_iccv2019, zhou_cvpr2020} and single-stage methods~\cite{liao_cvpr2020, wang_cvpr2020, kim_bumsoo_eccv2020}. 
In the two-stage methods, humans and objects are first individually localized by off-the-shelf object detectors, and then the region features from the localized area are used to predict action classes.
To incorporate contextual information, auxiliary features such as the features from the union region of a human and object bounding box, and locations of the bounding boxes in an image are often utilized. 
The single-stage methods predict interactions using the features of a heuristically-defined position such as a midpoint between a human and object center~\cite{liao_cvpr2020}.

While both two- and single-stage methods have shown significant improvement, 
they often suffer from errors attributed to the nature of convolutional neural networks (CNNs) and the heuristic way of using CNN features.
Figure~\ref{fig:fig1_heatmap} shows typical failure cases of conventional methods. 
In Fig.~\ref{fig:fig1_heatmap_a}, we can easily recognize from an entire image that a boy is washing a car.
It is difficult, however, for two-stage methods to predict the action class ``wash" since they typically use only the cropped bounding-box regions.
The regions sometimes miss contextually important cues located outside the human and object bounding box such as the hose in Fig.~\ref{fig:fig1_heatmap_a}.
Even though the features of union regions may contain such cues, these regions are frequently dominated by disturbing contents such as background and irrelevant humans and objects.
Figure~\ref{fig:fig1_heatmap_b} shows an example where multiple HOI instances are overlapped.
In such a case, CNN-based feature extractors are forced to capture features of both instances in the overlapped region, ending up in obtaining contaminated features.
The detection based on the contaminated features easily results in failures.
The single-stage methods attempt to capture the contextual information by pairing a target human and object from an early stage in feature extraction and extracting integrated features rather than individually treating the targets. 
To determine the regions from which integrated features are extracted, they rely on heuristically-designed location-of-interest such as a midpoint between a human and object center~\cite{liao_cvpr2020}.
However, such reliance sometimes causes a problem.
Fig.~\ref{fig:fig1_heatmap_c} shows an example where a target human and object are located distantly.
%%%%%%%%%%%%%%%%%%%%%%%%%%%%%%%%%%%%%%%%%%%%%%%%%%%%
In this example, the midpoint is located close to the man in the middle, who is not relevant to the target HOI instance.
Therefore, it is difficult to detect the target on the basis of the features around the midpoint.
Fig.~\ref{fig:fig1_heatmap_d} is an example where the midpoints of multiple HOI instances are close to each other.
In this case, CNN-based methods tend to make mis-detection due to the same reason as the one for the failure in Fig.~\ref{fig:fig1_heatmap_b}, \ie contaminated features.

To overcome these drawbacks, we propose QPIC, a query-based HOI detector that detects a human and object in a pairwise manner with image-wide contextual information. 
QPIC has a transformer~\cite{vaswani_nips2017} as a key component.
The attention mechanism used in QPIC scans through the entire area of an image and is expected to selectively aggregate contextually important information according to the contents of an image.
Moreover, we design QPIC's queries so that each query captures at most one human-object pair.
This enables to separately extract features of multiple HOI instances without contaminating them even when the instances are located closely.
These key designs of the attention mechanism and query-based pairwise detection make QPIC robust even under the difficult conditions such as the case where contextually important information appears outside the human and object bounding box (Fig.~\ref{fig:fig1_heatmap_a}), the target human and object are located distantly (Fig.~\ref{fig:fig1_heatmap_c}), and multiple instances are close to each other (Fig.~\ref{fig:fig1_heatmap_b} and~\ref{fig:fig1_heatmap_d}).
The key designs produce so effective embeddings that the subsequent detection heads may be fairly simple and intuitive.  

To summarize, our contributions are three-fold: 
(1) We propose a simple yet effective query-based HOI detector, QPIC, which incorporates contextually important information aggregated image-wide. To the best of our knowledge, this is the first work to introduce an attention- and query-based method to HOI detection. 
(2) We achieve significantly better performance than state-of-the-art methods on two challenging HOI detection benchmarks. 
(3) We conduct detailed analysis on the behavior of QPIC in relation to that of conventional methods, and reveal some of the important characteristics of HOI detection tasks that conventional methods could not capture but QPIC does relatively well.

\section{Related Work}
Two-stage HOI detection methods~\cite{chao_wacv2018, gao_eccv2020, gao_bmvc2018, gkioxari_cvpr2018, gupta_iccv2019, zhi_eccv2020, kim_dong_eccv2020, li_cvpr2020, li_cvpr2019, lin_ijcai2020, liu_eccv2020, ulutan_cvpr2020, wan_iccv2019, hai_eccv2020, xu_tmm2020, yang_ijcai2020, zhong_eccv2020, zhou_iccv2019, qi_eccv2018, wang_iccv2019, zhou_cvpr2020} utilize Faster R-CNN~\cite{ren_nips2015} or Mask R-CNN~\cite{he_iccv2017} to localize targets. Then, they crop features of backbone networks inside the localized regions. The cropped features are typically processed with multi-stream networks. Each stream processes features of target humans, those of objects, and some auxiliary features such as spatial configurations of the targets, and human poses either alone or in combination. Some of the two-stage methods~\cite{qi_eccv2018, ulutan_cvpr2020, hai_eccv2020, yang_ijcai2020, zhou_iccv2019} utilize graph neural networks to refine the features. These methods mainly focus on the second stage architecture, which uses cropped features to predict action classes. However, the cropped features sometimes lack contextual information outside the cropped regions or are contaminated by features of irrelevant targets, which results in the degradation of the performance.

Recently, single-stage methods~\cite{liao_cvpr2020, wang_cvpr2020, kim_bumsoo_eccv2020} that utilize integrated features from a pair of a human and object have been proposed to solve the problem in the individually cropped features.
Liao \etal~\cite{liao_cvpr2020} and Wang \etal~\cite{wang_cvpr2020} proposed a point-based interaction detection method that utilizes CenterNet~\cite{zhou_center_arxiv2019} as a base detector.
This method predicts action classes using integrated features collected at a midpoint between a human and object center.
In particular, Liao \etal's PPDM~\cite{liao_cvpr2020} achieves simultaneous object and interaction detection training, which is the most similar to our training approach. Kim \etal~\cite{kim_bumsoo_eccv2020} proposed UnionDet, which predicts the union bounding box of a human-object pair to extract integrated features. Although these methods attempt to capture contextual information by integrated features, they are still insufficient and sometimes contaminated due to the CNN's locality and heuristically-designed location-of-interests.

Our method differs from conventional methods in that we leverage a transformer to aggregate image-wide contextual features in a pairwise manner. We use DETR~\cite{carion_eccv2020} as a base detector and extend it for HOI detection. 

%%%%%%%%%%%%%%%%%%%%%%%%%%%%%%%%%%%%%%%%%%%%%%%%%%%%%%%%%%%%%%%%%%%%%%%%%%%%%%%%%%%%%%%%%%%%%%%%%%%%%%%%%%%%%%
\section{Proposed Method}
\begin{figure*}[t]
\centering
\includegraphics[keepaspectratio,width=0.9\linewidth]{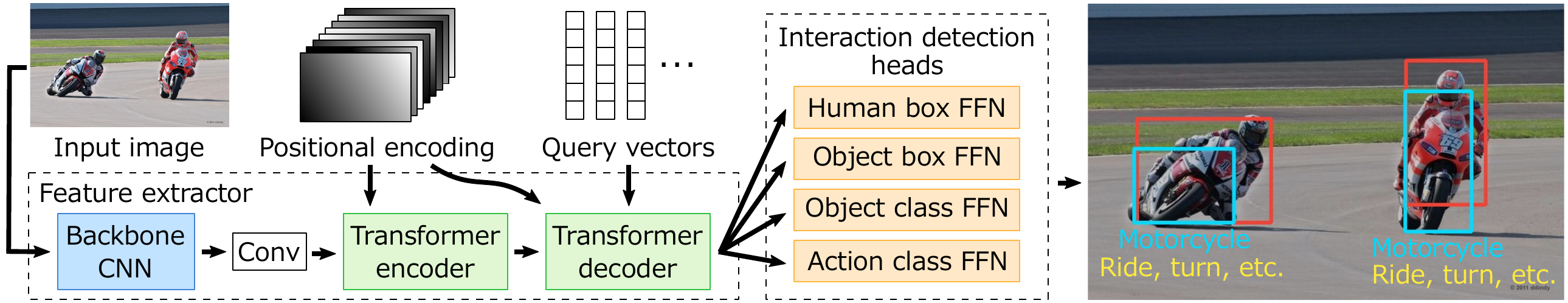}
\caption{Overall architecture of the proposed QPIC.}\label{fig:fig2_arch}
\vspace{-2.0ex}
\end{figure*}

To effectively extract important features for each HOI instance taking image-wide contexts into account, we propose to leverage transformer-based architecture as a base feature extractor.
We first explain the overall architecture in Sec.~\ref{subsec:architecture} and show that the detection heads following the base feature extractor can be simplified due to the rich features obtained in the base feature extractor.
In Sec.~\ref{subsec:int_train}, we show the concrete formulation of the loss function involved in the training.
Finally we explain how to use our method to detect HOI instances given a new image in Sec.~\ref{subsec:inference}.

%%%%%%%%%%%%%%%%%%%%%%%%%%%%%%%%%%%%%%%%%%%%%%%%%%%%%%%%%%%%%%%%%%%%%%%%%%%%%%%%%%%%%%%%%%%%%%%%%%%%%%%%%%%%%%
\subsection{Overall Architecture}\label{subsec:architecture}

Figure~\ref{fig:fig2_arch} illustrates the overall architecture of QPIC.
Given an input image $\bm{x} \in \mathbb{R}^{3 \times H \times W}$, a feature map $\bm{z_b} \in \mathbb{R}^{D_{b} \times H' \times W'}$ is calculated by an arbitrary off-the-shelf backbone network, where $H$ and $W$ are the height and width of the input image, $H'$ and $W'$ are those of the output feature map, and $D_{b}$ is the number of channels. 
Typically $H' < H$ and $W' < W$.
$\bm{z_b}$ is then input to a projection convolution layer with a kernel size of $1 \times 1$ to reduce the dimension from $D_b$ to $D_c$.

The transformer encoder takes this feature map with the reduced dimension $\bm{z_c} \in \mathbb{R}^{D_{c} \times H' \times W'}$ to produce another feature map with richer contextual information on the basis of the self-attention mechanism.
A fixed positional encoding $\bm{p} \in \mathbb{R}^{D_{c} \times H' \times W'}$~\cite{bello_iccv2019, parmar_icml2018, carion_eccv2020} is additionally input to the encoder to supplement the positional information, which the self-attention mechanism alone cannot inherently incorporate.
The encoded feature map $\bm{z_e} \in \mathbb{R}^{D_{c} \times H' \times W'}$ is then obtained as
$\bm{z_e} = f_{enc}\left(\bm{z_c}, \bm{p}\right)$, where $f_{enc}\left(\cdot, \cdot \right)$ is a set of stacked transformer encoder layers.

The transformer decoder transforms a set of learnable query vectors $\bm{Q} = \{\bm{q}_{i} | \bm{q}_{i} \in \mathbb{R}^{D_{c}}\}_{i=1}^{N_{q}}$ into a set of embeddings $\bm{D} = \{\bm{d}_{i} | \bm{d}_{i} \in \mathbb{R}^{D_{c}}\}_{i=1}^{N_{q}}$ that contain image-wide contextual information for HOI detection, referring to the encoded feature map $\bm{z_e}$ using the attention mechanism.
$N_q$ is the number of query vectors.
The queries are designed in such a way that one query captures at most one human-object pair and an interaction(s) between them.
$N_q$ is therefore set to be large enough so that it is always larger than the number of actual human-object pairs in an image.
The decoded embeddings are then obtained as 
$\bm{D} = f_{dec}\left(\bm{z_e}, \bm{p}, \bm{Q}\right)$,
where $f_{dec}\left(\cdot, \cdot, \cdot \right)$ is a set of stacked transformer decoder layers.
We use a positional encoding $\bm{p}$ again to incorporate the spatial information.

The subsequent interaction detection heads further processes the decoded embeddings to produce $N_q$ prediction results.
Here, we note that one or more HOIs corresponding to a human-object pair are mathematically defined by the following four vectors:
 a human-bounding-box vector normalized by the corresponding image size $\bm{b}^{(h)} \in [0, 1]^4$, 
 a normalized object-bounding-box vector $\bm{b}^{(o)} \in [0, 1]^4$, 
 an object-class one-hot vector $\bm{c} \in \{0, 1\}^{N_{obj}}$, where $N_{obj}$ is the number of object classes, 
 and an action-class vector $\bm{a} \in \{0, 1\}^{N_{act}}$, where $N_{act}$ is the number of action classes.
Note that $\bm{a}$ is not necessarily a one-hot vector because there may be multiple actions that correspond to a human-object pair.
Our interaction detection heads are composed of four small feed-forward networks (FFNs):
 human-bounding-box FFN $f_{h}$,
 object-bounding-box FFN $f_{o}$,
 object-class FFN $f_{c}$,
 and action-class FFN $f_{a}$, each of which is dedicated to predict one of the aforementioned 4 vectors, respectively.
This design of the interaction detection heads is fairly intuitive and simple compared with a number of state-of-the-art methods such as the point-detection and point-matching branch in PPDM~\cite{liao_cvpr2020} and the human, object, and spatial-semantic stream in DRG~\cite{gao_eccv2020}.
Thanks to the powerful embeddings that contain image-wide contextual information, QPIC does not have to rely on a rather complicated and heuristic design to produce the prediction.
One thing to note is that 
unlike many existing methods~\cite{chao_wacv2018, gao_eccv2020, gao_bmvc2018, gkioxari_cvpr2018, gupta_iccv2019, zhi_eccv2020, kim_dong_eccv2020, li_cvpr2020, li_cvpr2019, lin_ijcai2020, liu_eccv2020, ulutan_cvpr2020, wan_iccv2019, hai_eccv2020, xu_tmm2020, yang_ijcai2020, zhong_eccv2020, zhou_iccv2019, qi_eccv2018, wang_iccv2019, zhou_cvpr2020}, which first attempt to detect humans and objects individually and later pair them to find interactions, it is crucial to design queries in such a way that one query directly captures a human and object as a pair to more effectively extract features for interactions.
We will experimentally verify this claim in Sec.~\ref{subsec:head}.

The prediction of normalized human bounding boxes 
$\{\bm{\hat{b}}^{(h)}_i | \bm{\hat{b}}^{(h)}_i \in [0, 1]^4\}_{i=1}^{N_q}$, 
 that of object bounding boxes $\{\bm{\hat{b}}^{(o)}_i | \bm{\hat{b}}^{(o)}_i \in [0, 1]^4\}_{i=1}^{N_q}$, 
 the probability of object classes $\{\bm{\hat{c}}_i | \bm{\hat{c}}_i \in [0, 1]^{N_{obj} + 1}, \sum_{j=1}^{N_{obj} + 1}{\bm{\hat{c}}_i}(j)=1\}_{i=1}^{N_q}$, where $\bm{v}(j)$ denotes the $j$-th element of $\bm{v}$,
 and the probability of action classes $\{\bm{\hat{a}}_i | \bm{\hat{a}}_i \in [0, 1]^{N_{act}}\}_{i=1}^{N_q}$, 
are calculated as 
$\bm{\hat{b}}^{(h)}_{i} = \sigma\left(f_{h}\left(\bm{d}_{i}\right)\right), 
\bm{\hat{b}}^{(o)}_{i} = \sigma\left(f_{o}\left(\bm{d}_{i}\right)\right), 
\bm{\hat{c}}_{i} = \varsigma\left(f_{c}\left(\bm{d}_{i}\right)\right), 
\bm{\hat{a}}_{i} = \sigma\left(f_{a}\left(\bm{d}_{i}\right)\right)
$
, respectively.
$\sigma, \varsigma$ are the sigmoid and softmax functions, respectively. 
Note that $\bm{\hat{c}}_{i}$ has the $(N_{obj}+1)$-th element to indicate that the $i$-th query has no corresponding human-object pair, while an additional element of $\bm{\hat{a}}_{i}$ to indicate ``no action'' is not necessary because we use the sigmoid function rather than the softmax function to calculate the action-class probabilities for co-occuring actions. 

%%%%%%%%%%%%%%%%%%%%%%%%%%%%%%%%%%%%%%%%%%%%%%%%%%%%%%%%%%%%%%%%%%%%%%%%%%%%%%%%%%%%%%%%%%%%%%%%%%%%%%%%%%%%%%
\subsection{Loss Calculation}\label{subsec:int_train}
The loss calculation is composed of two stages: the bipartite matching stage between predictions and ground truths, and the loss calculation stage for the matched pairs.

For the bipartite matching, 
we follow the training procedure of DETR~\cite{carion_eccv2020} and use the Hungarian algorithm~\cite{kuhn_1955}. 
Note that this design obviates the process of suppressing over-detection as described in~\cite{carion_eccv2020}.
We first pad the ground-truth set of human-object pairs with $\phi$ (no pairs) so that the ground-truth set size becomes $N_q$.
We then leverage the Hungarian algorithm to determine the optimal assignment $\hat{\omega}$ among the set of all possible permutations of $N_q$ elements $\bm{\Omega}_{N_q}$,
 i.e. $\hat{\omega} = \argmin_{\omega \in \bm{\Omega}_{N_q}}{\sum_{i=1}^{N_q}{\mathcal{H}_{i,\omega(i)}}}$, where $\mathcal{H}_{i,j}$ is the matching cost for the pair of $i$-th ground truth and $j$-th prediction.
The matching cost $\mathcal{H}_{i, j}$ consists of four types of costs: 
the box-regression cost $\mathcal{H}^{(b)}_{i, j}$, intersection-over-union (IoU) cost $\mathcal{H}^{(u)}_{i, j}$, object-class cost $\mathcal{H}^{(c)}_{i, j}$, and action-class cost $\mathcal{H}^{(a)}_{i, j}$.
Denoting $i$-th ground truth for
 the normalized human bounding box by $\bm{b}_{i}^{(h)} \in [0, 1]^4$, 
 normalized object bounding box by $\bm{b}_{i}^{(o)} \in [0, 1]^4$, 
 object-class one-hot vector by $\bm{c}_{i} \in \{0, 1\}^{N_{obj}}$,
 and action class by $\bm{a}_{i} \in \{0, 1\}^{N_{act}}$,
 the aforementioned costs are formulated as follows.
\begin{align}
    \mathcal{H}_{i, j} ={} & \mathbbm{1}_{\{i \not\in \bm{\Phi}\}}\left[\eta_{b} \mathcal{H}^{(b)}_{i, j} + \eta_{u} \mathcal{H}^{(u)}_{i, j} + \eta_{c} \mathcal{H}^{(c)}_{i, j} + \eta_{a} \mathcal{H}^{(a)}_{i, j}\right], \\
    \mathcal{H}^{(b)}_{i, j} ={} & \max\left\{\left\|\bm{b}^{(h)}_{i} - \bm{\hat{b}}^{(h)}_{j}\right\|_{1}, \left\|\bm{b}^{(o)}_{i} - \bm{\hat{b}}^{(o)}_{j}\right\|_{1}\right\},\label{eq:cost_box} \\
    \nonumber\mathcal{H}^{(u)}_{i, j} ={} & \max\left\{-GIoU\left(\bm{b}^{(h)}_{i}, \bm{\hat{b}}^{(h)}_{j}\right),\right.\\
        &{} \qquad\qquad \left. -GIoU\left(\bm{b}^{(o)}_{i}, \bm{\hat{b}}^{(o)}_{j}\right)\right\},\label{eq:cost_iou} \\
    \mathcal{H}^{(c)}_{i, j} ={} & -\bm{\hat{c}}_j(k)\quad s.t.\quad\bm{c}_i(k)=1, \\
    \mathcal{H}^{(a)}_{i, j} ={} & -\frac{1}{2}\left(
    \frac{\bm{a}^{\intercal}_{i}\bm{\hat{a}}_{j}}
        {\left\|\bm{a}_{i}\right\|_{1} + \epsilon}
    + \frac{\left(\bm{1} - \bm{a}_{i}\right)^{\intercal}\left(\bm{1} - \bm{\hat{a}}_{j}\right)}
        {\left\|\bm{1} - \bm{a}_{i}\right\|_{1} + \epsilon}
    \right),\label{eq:cost_action} 
\end{align}
where $\bm{\Phi}$ is a set of ground-truth indices that correspond to $\phi$, 
$GIoU\left(\cdot, \cdot \right)$ is the generalized IoU~\cite{rezatofighi_cvpr2019},
$\epsilon$ is a small positive value introduced to avoid zero divide,
and $\eta_b$, $\eta_u$, $\eta_c$, and $\eta_a$ are the hyper-parameters.
We use two types of bounding-box cost $\mathcal{H}^{(b)}_{i, j}$ and $\mathcal{H}^{(u)}_{i, j}$ following~\cite{carion_eccv2020}.
In calculating $\mathcal{H}^{(b)}_{i, j}$ and $\mathcal{H}^{(u)}_{i, j}$, instead of minimizing the average of a human and object-bounding-box cost, we minimize the larger of the two to prevent the matching from being undesirably biased to either if one cost is significantly lower than the other. 
We design $\mathcal{H}^{(a)}_{i, j}$ so that the costs of both positive and negative action classes are taken into account. 
In addition, we formulate it using the weighted average of the two with the inverse number of nonzero elements as the weights rather than using the vanilla average.
This is necessary to balance the effect from the two costs because the number of positive action classes is typically much smaller than that of negative action classes.

The loss to be minimized in the training phase is calculated on the basis of the matched pairs
as follows.
\begin{align}
    \mathcal{L} ={} & \lambda_{b} \mathcal{L}_{b} + \lambda_{u} \mathcal{L}_{u} + \lambda_{c} \mathcal{L}_{c} + \lambda_{a} \mathcal{L}_{a}, \\
    \nonumber\mathcal{L}_{b} ={} & \frac{1}{|\bar{\bm{\Phi}}|} \sum_{i=1}^{N_{q}} \mathbbm{1}_{\{i \not\in \bm{\Phi}\}}\left[
        \left\|\bm{b}^{(h)}_i - \bm{\hat{b}}^{(h)}_{\hat{\omega}\left(i\right)}\right\|_{1} \right.\\
        &{} \qquad\qquad\qquad\qquad \left. + \left\|\bm{b}^{(o)}_i - \bm{\hat{b}}^{(o)}_{\hat{\omega}\left(i\right)}\right\|_{1}\right], \\
    \nonumber\mathcal{L}_{u} ={} & \frac{1}{|\bar{\bm{\Phi}}|} \sum_{i=1}^{N_{q}} \mathbbm{1}_{\{i \not\in \bm{\Phi}\}}\left[2 - 
        GIoU\left(\bm{b}^{(h)}_i, \bm{\hat{b}}^{(h)}_{\hat{\omega}\left(i\right)}\right) \right.\\
        &{} \qquad\qquad\qquad\qquad \left. - GIoU\left(\bm{b}^{(o)}_i, \bm{\hat{b}}^{(o)}_{\hat{\omega}\left(i\right)}\right)\right], \\
    \nonumber\mathcal{L}_{c} ={} & \frac{1}{N_{q}}\sum_{i=1}^{N_{q}}\left\{
        \mathbbm{1}_{\{i \not\in \bm{\Phi}\}}\left[-\log\bm{\hat{c}_{\hat{\omega}\left(i\right)}}(k)\right]\right.\\
        \nonumber&{} \qquad\quad \left. + \mathbbm{1}_{\{i \in \bm{\Phi}\}}\left[-\log\bm{\hat{c}_{\hat{\omega}\left(i\right)}}(N_{obj}+1)\right]\right\}\\
        &{} \qquad\qquad\quad s.t.\quad\bm{c}_i(k)=1, \\
    \nonumber\mathcal{L}_{a} ={} &  \frac{1}{\sum_{i=1}^{N_{q}}\mathbbm{1}_{\{i \not\in \bm{\Phi}\}}\left\|\bm{a}_i\right\|_{1}}\sum_{i=1}^{N_{q}}\left\{
        \mathbbm{1}_{\{i \not\in \bm{\Phi}\}}\left[l_{f}\left(\bm{a}_i, \bm{\hat{a}_{\hat{\omega}\left(i\right)}}\right)\right]\right.\\
        &{} \qquad\qquad \left. + \mathbbm{1}_{\{i \in \bm{\Phi}\}}\left[l_{f}\left(\bm{0}, \bm{\hat{a}_{\hat{\omega}\left(i\right)}}\right)\right]
        \right\},
\end{align}
where $\lambda_{b}$,  $\lambda_{u}$, $\lambda_{c}$ and $\lambda_{a}$ are the hyper-parameters for adjusting the weights of each loss, and $l_{f}\left(\cdot, \cdot\right)$ is the element-wise focal loss function~\cite{lin_iccv2017}. 
For the hyper-parameters of the focal loss, we use the default settings described in~\cite{zhou_center_arxiv2019}. 
%%%%%%%%%%%%%%%%%%%%%%%%%%%%%%%%%%%%%%%%%%%%%%%%%%%%%%%%%%%%%%%%%%%%%%%%%%%%%%%%%%%%%%%%%%%%%%%%%%%%%%%%%%%%%%
\subsection{Inference for Interaction Detection}\label{subsec:inference}
As previously mentioned, the detection result of an HOI is represented by the following four components, $\langle${\it human bounding box, object bounding box, object class, action class}$\rangle$.
Our interaction detection heads are designed so intuitively that all we need to do is to pick up the corresponding information from each head. 
Formally, we set the prediction results corresponding to the $i$-th query and $j$-th action as $\langle\bm{\hat{b}^{(h)}_{i}}, \bm{\hat{b}^{(o)}_{i}}, \argmax_{k}{\bm{\hat{c}_{i}}(k)}, j\rangle$.
We define a score of the HOI instance as 
$\left\{\max_{k}{\bm{\hat{c}_{i}}(k)}\right\}\bm{\hat{a}_{i}}(j)$,
and regard this instance to be present if the score is higher than a threshold.

\section{Experiments}
\subsection{Datasets and Evaluation Metrics}
We conducted extensive experiments on two HOI detection datasets: HICO-DET~\cite{chao_wacv2018} and V-COCO~\cite{gupta_arxiv2015}. 
We followed the standard evaluation scheme.
HICO-DET contains 38,118 and 9,658 images for training and testing, respectively. 
The images are annotated with 80 object and 117 action classes.
V-COCO, which originates from the COCO dataset, contains 2,533, 2,867, and 4,946 images for training, validation, and testing, respectively. 
The images are annotated with 80 object and 29 action classes. 

For the evaluation metrics, we use the mean average precision (mAP). 
A detection result is judged as a true positive if the predicted human and object bounding box have IoUs larger than 0.5 with the corresponding ground-truth bounding boxes, and the predicted action class is correct. In the HICO-DET evaluation, the object class is also taken into account for the judgment. The AP is calculated per object and action class pair in the HICO-DET evaluation, while that is calculated per action class in the V-COCO evaluation.

For HICO-DET, we evaluate the performance in two different settings following~\cite{chao_wacv2018}: {\it default setting} and {\it known-object setting}.
In the former setting, APs are calculated on the basis of all the test images, while in the latter setting, each AP is calculated only on the basis of images that contain the object class corresponding to each AP. 
In each setting, we report the mAP over three set types: a set of 600 HOI classes ({\it full}), a set of 138 HOI classes that have less than 10 training instances ({\it rare}), and a set of 462 HOI classes that have 10 or more training instances ({\it non-rare}).
Unless otherwise stated, we use the default full setting in the analysis.
In V-COCO, a number of HOIs are defined with no object labels.
To deal with this situation, we evaluate the performance in two different scenarios following V-COCO's official evaluation scheme. 
In scenario 1, detectors are required to report cases in which there is no object, while in scenario 2, we just ignore the prediction of an object bounding box in these cases.

\subsection{Implementation Details}
We use ResNet-50 and ResNet-101~\cite{he_cvpr2016} as a backbone feature extractor. 
Both transformer encoder and decoder consist of 6 transformer layers with a multi-head attention of 8 heads. The reduced dimension size $D_c$ is set to 256, and the number of query vectors $N_q$ is set to 100.
The human- and object-bounding-box FFNs have 3 linear layers with ReLU activations, while the object- and action-class FFNs have 1 linear layer.

For training QPIC, we initialize the network with the parameters of DETR~\cite{carion_eccv2020} trained with the COCO dataset. Note that for the V-COCO training, we exclude the COCO's training images that are contained in the V-COCO test set when pre-training DETR\footnote{A few previous works inappropriately use COCO train2017 set for pre-training, whose images are contained in the V-COCO test set.}. 
QPIC is trained for 150 epochs using the AdamW~\cite{loshchiloy_iclr2019} optimizer with the batch size 16, initial learning rate of the backbone network $10^{-5}$, that of the others $10^{-4}$, and the weight decay $10^{-4}$. Both learning rates are decayed after 100 epochs. The hyper-parameters for the Hungarian costs $\eta_b, \eta_u, \eta_c,$ and $\eta_a$, and those for the loss weights $\lambda_b, \lambda_u, \lambda_c,$ and $\lambda_a$ are set to 2.5, 1, 1, 1, 2.5, 1, 1, and 1, respectively.
Following~\cite{liao_cvpr2020}, we select 100 high scored detection results from all the predictions for fair comparison.
Please see the supplementary material for more details.

\subsection{Comparison to State-of-the-Art}\label{subsec:sota}
\begin{table}[t]
    \caption{Comparison against state-of-the-art methods on HICO-DET. The top, middle, and bottom blocks show the mAPs of the two-stage, single-stage, and our methods, respectively.}
    \label{table:comp_hico}
    \centering
    \small
    \setlength{\tabcolsep}{2pt}
    \begin{tabular}{@{}lcccccc@{}}
        \toprule
        & \multicolumn{3}{c}{Default} & \multicolumn{3}{c}{Known object} \\
        \cmidrule(lr){2-4}\cmidrule(lr){5-7}
        Method & full & rare & non-rare & full & rare & non-rare \\
        \midrule
        FCMNet~\cite{liu_eccv2020} & 20.41 & 17.34 & 21.56 & 22.04 & 18.97 & 23.13 \\
        ACP~\cite{kim_dong_eccv2020} & 20.59 & 15.92 & 21.98 & -- & -- & -- \\
        VCL~\cite{zhi_eccv2020} & 23.63 & 17.21 & 25.55 & 25.98 & 19.12 & 28.03 \\
        DRG~\cite{gao_eccv2020} & 24.53 & 19.47 & 26.04 & 27.98 & 23.11 & 29.43 \\
        \midrule
        UnionDet~\cite{kim_bumsoo_eccv2020} & 17.58 & 11.72 & 19.33 & 19.76 & 14.68 & 21.27 \\
        Wang \etal~\cite{wang_cvpr2020} & 19.56 & 12.79 & 21.58 & 22.05 & 15.77 & 23.92 \\
        PPDM~\cite{liao_cvpr2020} & 21.73 & 13.78 & 24.10 & 24.58 & 16.65 & 26.84 \\
        \midrule
        Ours (ResNet-50) & 29.07 & 21.85 & 31.23 & 31.68 & 24.14 & 33.93 \\
        Ours (ResNet-101) & \textbf{29.90} & \textbf{23.92} & \textbf{31.69} & \textbf{32.38} & \textbf{26.06} & \textbf{34.27} \\
        \bottomrule
    \end{tabular}
    \vspace{-1.0ex}
\end{table}

\begin{table}[t]
    \caption{Comparison against state-of-the-art methods on V-COCO. The split of the blocks are the same as Table~\ref{table:comp_hico}.}
    \label{table:comp_vcoco}
    \centering
    \small
    \begin{tabular}{@{}lcc@{}}
        \toprule
        Method & Scenario 1 & Scenario 2 \\
        \midrule
        VCL~\cite{zhi_eccv2020} & 48.3 & -- \\
        DRG~\cite{gao_eccv2020} & 51.0 & -- \\
        ACP~\cite{kim_dong_eccv2020} & 53.0 & -- \\
        FCMNet~\cite{liu_eccv2020} & 53.1 & -- \\
        \midrule
        UnionDet~\cite{kim_bumsoo_eccv2020} & 47.5 & 56.2 \\
        Wang \etal~\cite{wang_cvpr2020} & 51.0 & -- \\
        \midrule
        Ours (ResNet-50) & \textbf{58.8} & \textbf{61.0} \\
        Ours (ResNet-101) & 58.3 & 60.7\\
        \bottomrule
    \end{tabular}
    \vspace{-2.0ex}
\end{table}

\begin{figure*}[t]
\centering
\subfloat[AP depending on the distance between a human and object center.]{%
    \label{fig:fig5_distance_analysis}
    \includegraphics[clip,keepaspectratio,width=1.\columnwidth]{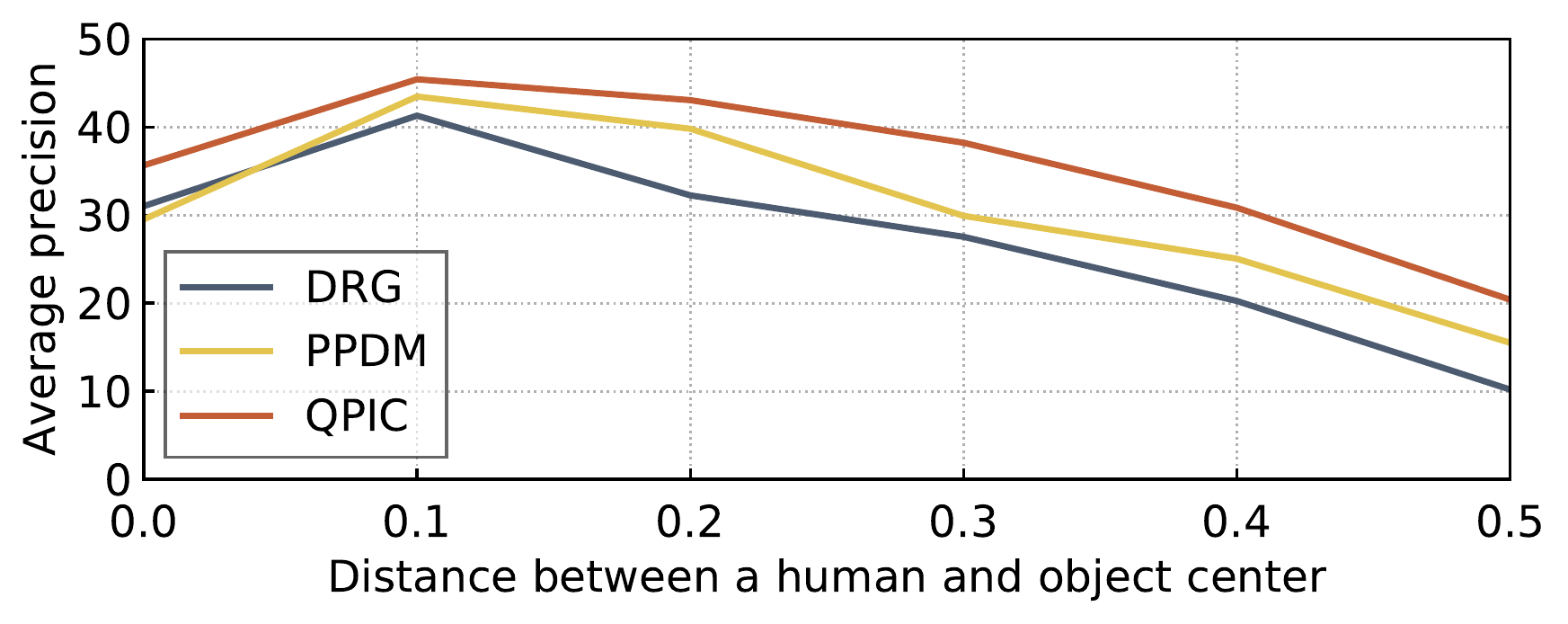}%
}
\subfloat[AP depending on the larger area of a human and object bounding box.]{%
    \label{fig:fig5_area_analysis}
    \includegraphics[clip,keepaspectratio,width=1.\columnwidth]{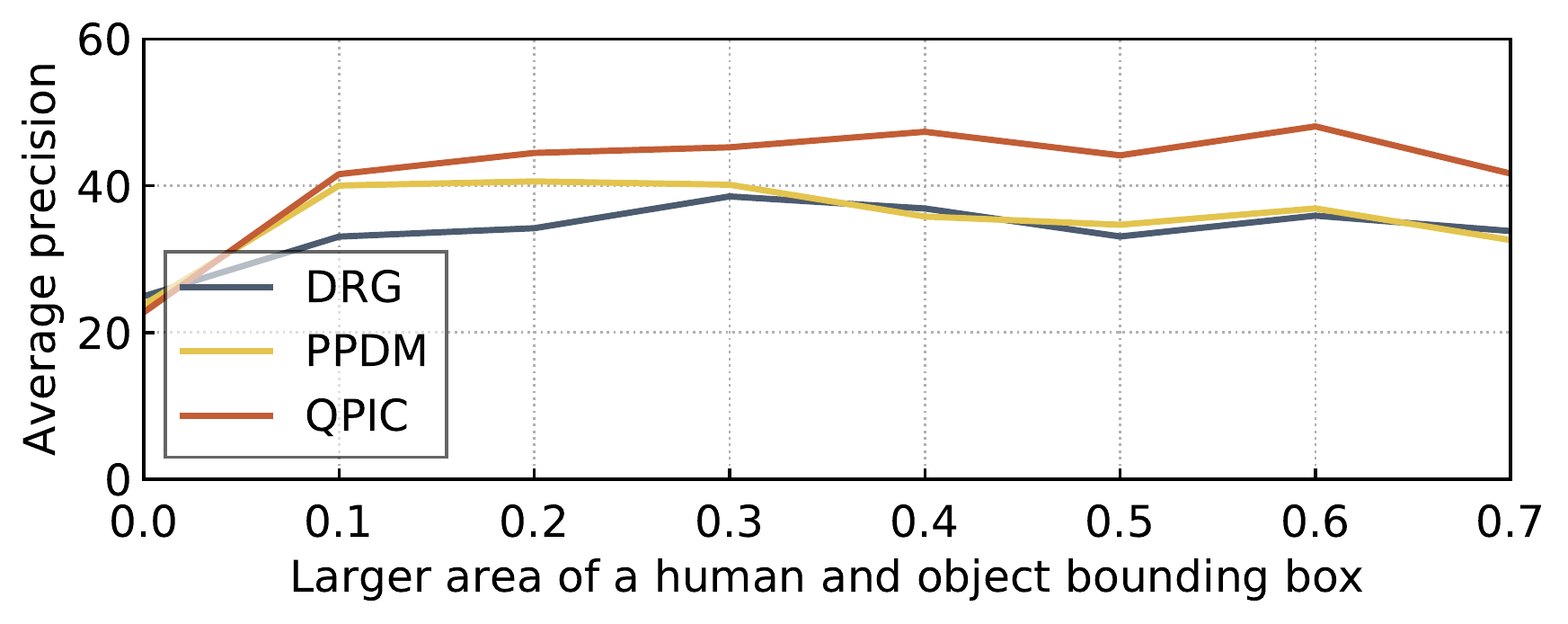}%
}
\caption{Performance analysis on different spatial distribution of HOIs evaluated on HICO-DET.}\label{fig:space_analysis}
\vspace{-2.0ex}
\end{figure*}

We first show the comparison of our QPIC with the latest HOI detection methods including both two- and single-stage methods in Table~\ref{table:comp_hico}. 
As seen from the table, QPIC outperforms both state-of-the-art two- and single-stage methods in all the settings. 
QPIC with the ResNet-101 backbone yields an especially significant gain of 5.37 mAP (relatively 21.9\%) compared with DRG~\cite{gao_eccv2020} and 8.17 mAP (37.6\%) compared with PPDM~\cite{liao_cvpr2020} in the default full setting. 
Table~\ref{table:comp_vcoco} shows the comparison results on V-COCO. QPIC achieves state-of-the-art performance among all the baseline methods. QPIC with the ResNet-50 backbone achieves a 5.7 mAP (10.7\%) gain over FCMNet~\cite{liu_eccv2020}, which is the strongest baseline. Unlike in the HICO-DET result, the ResNet-50 backbone shows better performance than the ResNet-101 backbone probably because the number of training samples in V-COCO is insufficient to train the large network. 
Overall, these comparison results demonstrate the dataset-invariant effectiveness of QPIC.

We then investigate in which cases QPIC especially achieves superior performance compared with the strong baselines.
To do so, we compare the performance of QPIC in detail with DRG~\cite{gao_eccv2020} and PPDM~\cite{liao_cvpr2020}, which are the strongest baselines of the two- and single-stage methods, respectively. 
We use the ResNet-50 backbone for QPIC in this comparison.
Note that hereinafter the distance and area are calculated in normalized image coordinates.
Figure~\ref{fig:fig5_distance_analysis} shows how the performances change as the distance between the center points of a paired human and object bounding box grows.
We split HOI instances into bins of size 0.1 according to the distances, and calculate the APs of each bin that has at least 1,000 HOI instances.
As shown in Fig.~\ref{fig:fig5_distance_analysis}, the relative gaps of the performance between QPIC and the other two methods become more evident as the distance grows.
The graph suggests three things; 
 HOI detection tends to become more difficult as the distance grows,
 the distant case is especially difficult for CNN-based methods,
 and QPIC relatively better deals with this difficulty.
The possible explanation for these results is that the features of the CNN-based methods, which rely on limited receptive fields for the feature aggregation, 
cannot include contextually important information or are dominated by irrelevant information in the distant cases, while the features of QPIC are more effective thanks to the ability of selectively extracting image-wide contextual information.
Figure~\ref{fig:fig5_area_analysis} presents how the performances change as the areas of target human and object bounding boxes grow. 
We pick up the larger area of a target human and object bounding box involved in each HOI instance.
We then split HOI instances into bins of size 0.1 according to the area, and calculate the APs of each bin that has at least 1,000 HOI instances.
As illustrated in Fig.~\ref{fig:fig5_area_analysis}, 
the gaps of the APs between the conventional methods and QPIC tend to grow as the area increases.
This is probably because of the combination of the following two reasons;
 if the area becomes bigger, the area tends to more often include harmful regions such as another HOI instance,
 and the conventional methods mix up the irrelevant features in such situation, whereas the attention mechanism and the query-based framework enable to selectively aggregate effective features in a separated manner for each HOI instance. 
These results reveal that the QPIC's significant improvement shown in Table~\ref{table:comp_hico} and Table~\ref{table:comp_vcoco} is likely to be brought by its nature of robustness to diverse spatial distribution of HOIs, probably originating from its capability of aggregating image-wide contextual features for each HOI instance. This observation is further confirmed qualitatively in Sec.~\ref{sec:quality}.

\subsection{Ablation Study}\label{subsec:quantitative}
To understand the key ingredients of QPIC's superiority shown in Sec.~\ref{subsec:sota}, we analyze the key building blocks one by one in detail. We first analyze the interaction detection heads in Sec.~\ref{subsec:head} and subsequently analyze the transformer-based feature extractor in Sec.~\ref{subsec:feature_extractor}.

\vspace{-2.0ex}

\subsubsection{Analysis on Detection Heads}\label{subsec:head}
\begin{figure*}[t]
\centering
\subfloat[Interaction detection with point matching.]{%
    \label{fig:fig3_variants_b}
    \includegraphics[clip,keepaspectratio,width=0.92\columnwidth]{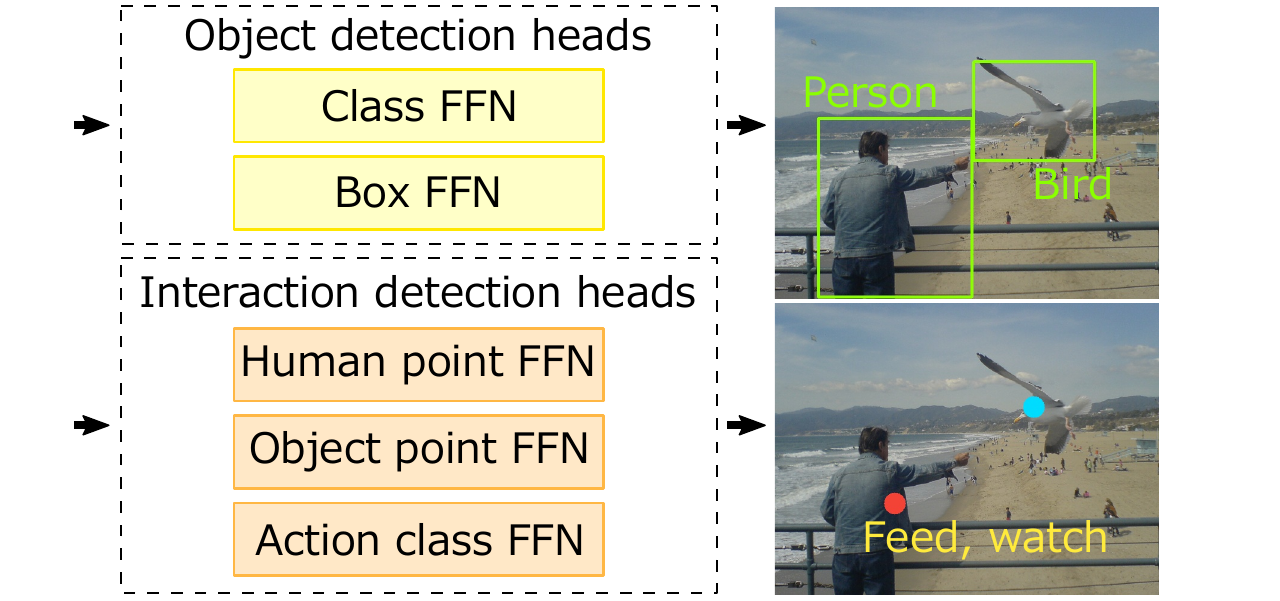}%
}
\subfloat[Interaction detection with two-stage like approach.]{%
    \label{fig:fig3_variants_a}
    \includegraphics[clip,keepaspectratio,width=1.\columnwidth]{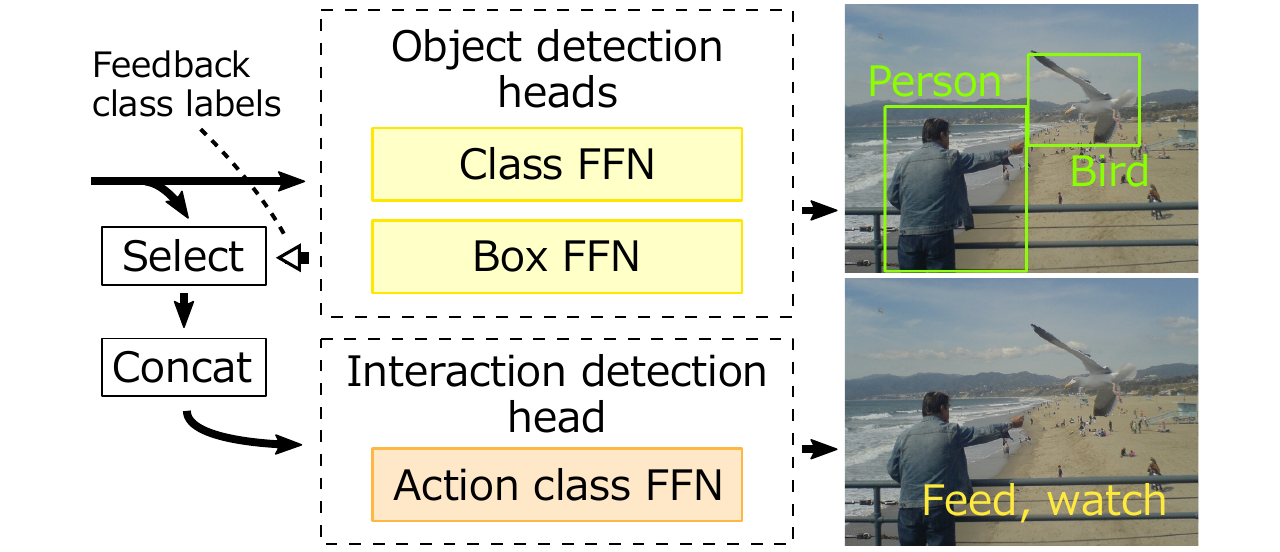}%
}
\caption{Implemented variants for analyzing detection heads. These heads are on top of our transformer-based feature extractor.}\label{fig:variants}
\vspace{-2.0ex}
\end{figure*}

\begin{table}[t]
    \caption{Evaluation results of the various detection heads.}
    \label{table:variants}
    \centering
    \small
    \setlength{\tabcolsep}{4pt}
    \begin{tabular}{@{}ccc@{}}
        \toprule
        Base method & Detection heads & HICO-DET (mAP) \\
        \midrule
        \multirow{3}{*}{\shortstack{Ours\\ (ResNet-50)}} & Simple (original) & 29.07 \\
        & Point matching (Fig.~\ref{fig:fig3_variants_b})  & 29.04 \\
        & Two-stage like (Fig.~\ref{fig:fig3_variants_a}) & 26.18 \\
        \midrule
        \multirow{2}{*}{{\shortstack{PPDM~\cite{liao_cvpr2020}\\ (Hourglass-104)}}} & Simple & 17.45 \\
        & Point matching (original) & 21.73 \\
        \bottomrule
    \end{tabular}
    \vspace{-2.0ex}
\end{table}

\paragraph{Feasibility of simple heads.}
As previously mentioned, the inference process of QPIC is simplified thanks to the enriched features from the transformer-based feature extractor. 
To confirm that this simple prediction is sufficient for QPIC, we investigated if the detection accuracy increases by leveraging a typical point-matching-based detection heads presented in~\cite{liao_cvpr2020}, which is one of the best performing heuristically-designed heads. 
Figure~\ref{fig:fig3_variants_b} represents the implemented heads. 
A notable difference from the original simple heads lies in that the interaction detection heads output center points of target humans and objects instead of bounding boxes. Consequently, the outputs from the interaction detection heads need to be fused with the outputs from the object detection heads with point matching. 
Note that in this implementation, duplicate detection results that share an identical human-object pair needs to be suppressed by some means such as non-maximum suppression. 
Table~\ref{table:variants} shows the evaluation results. 
As seen from Table~\ref{table:variants}, the point-matching-based heads exhibit no performance improvement over the simple heads, which indicates that the simple detection heads are enough and we do not have to manually design complicated detection heads.

\vspace{-2.0ex}

\paragraph{Importance of pairwise detection.}
Although the detection heads can be as simple as we present, we claim that there is a crucial aspect that must be covered in the design of the heads. It is to treat a target human and object as a pair from early stages rather than to first detect them individually and later integrate the features from the cropped regions corresponding to the detection, as typically done in two-stage approaches.
We assume the features from the cropped regions do not contain enough contextual information because sometimes the regions in an image other than a human and object bounding boxes play a crucial role in HOI detection (see Fig.~\ref{fig:fig1_heatmap_a} for example). 
We verify this claim by looking into the performance of the two-stage like detection-heads on top of our transformer-based feature extractor, which is exactly the same as original QPIC.
Figure~\ref{fig:fig3_variants_a} illustrates the implemented detection heads. 
This model first derives object detection results from the object detection heads.
Then, the results are used to create all the possible human-object pairs.
The features of each pair is constructed by concatenating the features from the human and object bounding boxes.
The interaction detection head predicts action classes of all the pairs on the basis of the concatenated features.
As seen from Table~\ref{table:variants}, two-stage like method yields worse performance than the original. 
This observation indicates that the two-stage methods, which rely on individual feature extraction, do not perform well even with our strong feature extractor, and suggests the importance of the pairwise feature extraction in heads for HOI detection. 

\vspace{-2.0ex}

\subsubsection{Analysis on Feature Extractor}\label{subsec:feature_extractor}
\paragraph{Importance of a transformer.}
To confirm that a transformer-based feature extractor is key to make the simple heads sufficiently work for HOI detection as discussed in Sec.~\ref{subsec:head}, we replace QPIC's transformer-based feature extractor by a CNN-based counterpart and examine how the performance changes.
We utilize the Hourglass-104 backbone used in PPDM~\cite{liao_cvpr2020} in this experiment.
Table~\ref{table:variants} shows the performance of the original point-matching-based PPDM as well as its simple-heads variant.
The simple-heads variant directly predicts all the information corresponding to a human-object pair on the basis of the features extracted in the feature-extraction stage, just as QPIC's simple heads do.
More concretely, not only a human point, an object point, and action classes, but also a human-bounding-box size, an object-bounding-box size, and an object class are directly predicted on the basis of the features at the midpoint between the human and object centers.
As Table~\ref{table:variants} shows, the simple-heads variant exhibits far worse performance than QPIC.
This implicates that the CNN-based feature extractor is not as powerful as our transformer-based feature extractor, so the simple heads cannot be leveraged with it.
In addition, we find that the point-matching-based heads, which is the original version of PPDM, achieve higher performance than the simple ones, implying that there is a room for increasing accuracy by heuristically designing the heads if the feature extractor is not so powerful, which is not the case with our powerful transformer-based feature extractor.

\vspace{-2.0ex}

\paragraph{Importance of a decoder.}
\begin{figure*}[t]
\centering
\subfloat[]{%
    \label{fig:fig4_qualitative_a}
    \includegraphics[clip,keepaspectratio,width=0.23\textwidth]{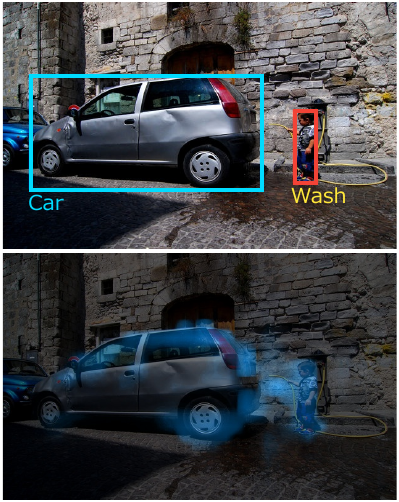}%
}
\subfloat[]{%
    \label{fig:fig4_qualitative_b}
    \includegraphics[clip,keepaspectratio,width=0.23\textwidth]{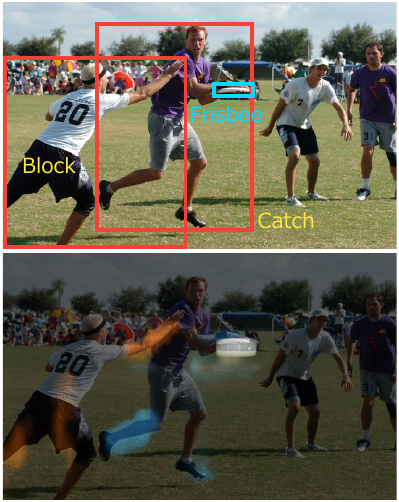}%
}
\subfloat[]{%
    \label{fig:fig4_qualitative_c}
    \includegraphics[clip,keepaspectratio,width=0.23\textwidth]{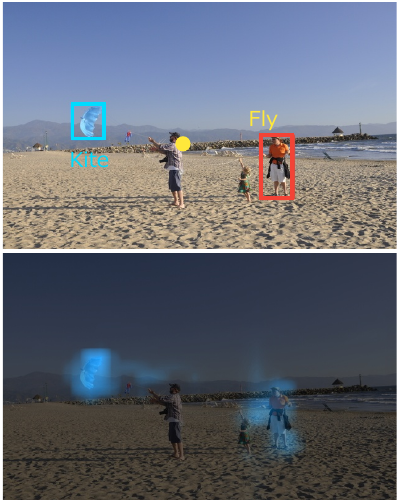}%
}
\subfloat[]{%
    \label{fig:fig4_qualitative_d}
    \includegraphics[clip,keepaspectratio,width=0.23\textwidth]{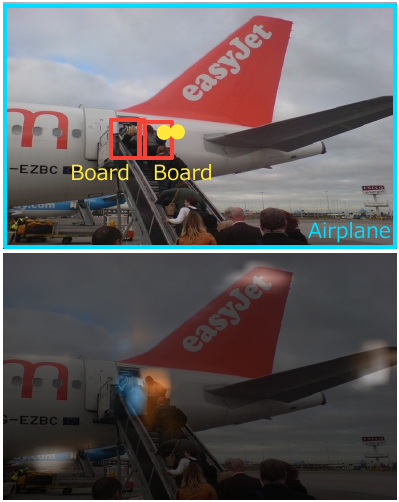}%
}
\caption{Failure cases of conventional detectors (top row, same as Fig.~\ref{fig:fig1_heatmap}) and attentions of QPIC (bottom row). In (b) and (d), the attentions corresponding to different HOI instances are drawn with blue and orange, and the areas where two attentions overlap are drawn with white.}
\label{fig:fig4_qualitative}
\vspace{-2.0ex}
\end{figure*}

\begin{table}[t]
    \caption{Effect of the transformer encoder and decoder.}
    \label{table:ablation}
    \centering
    \small
    \begin{tabular}{@{}cccc@{}}
        \toprule
        \multirow{2}{*}{\shortstack{Transformer\\ encoder}} & \multirow{2}{*}{\shortstack{Transformer\\ decoder}} &\multirow{2}{*}{\shortstack{HICO-DET\\ (mAP)}}&\multirow{2}{*}{\shortstack{COCO\\ (mAP)}} \\
        &&&\\
        \midrule
        && 18.89 & 34.6 \\
        \ding{51} && 20.07 & 35.1 \\
        &\ding{51}& 26.75 & 38.7 \\
        \ding{51} &\ding{51}& 29.27 & 43.5 \\
        \bottomrule
    \end{tabular}
    \vspace{-2.0ex}
\end{table}

To further dig into the transformer to find out the essential component for HOI detection, we compare four variants listed in Table~\ref{table:ablation}. 
The model without the decoder leverages the point-matching-based method like PPDM~\cite{liao_cvpr2020} on top of the encoder's output (with encoder) or on top of the base features (without encoder).
The model with the decoder utilizes the point-matching-based heads (Fig.~\ref{fig:fig3_variants_b}) for fair comparison.
We use the ResNet-101 backbone for all the variants.
As seen from Table~\ref{table:ablation}, the transformer encoder yields merely slight improvement on HICO-DET
(2.52 and 1.18 mAP with and without the decoder, respectively), 
while the decoder remarkably boosts the performance 
(9.20 and 7.86 mAP with and without the encoder, respectively). 
These results indicate that the decoder plays a vital role in HOI detection. 
Additionally, we evaluate the performance on COCO to compare the degrees of improvement in object detection and HOI detection. 
As seen in the table, the relative performance improvement brought by the decoder for object detection (on COCO) is 23.9\% and 11.8\% with and without the encoder, respectively, while that for HOI detection (on HICO-DET) is 45.8\% and 41.6\% with and without the encoder, respectively. 
This means that the decoder is more effective in an HOI detection task than in an object detection task.
This is probably because the regions of interest (ROI) are mostly consolidated in a single area in object detection tasks, while in HOI detection tasks, the ROI can be diversely distributed image-wide.
CNNs, which rely on localized receptive fields, can deal with the former case relatively easily, whereas the image-wide feature aggregation of the decoder is crutial for the latter case.
%%%%%%%%%%%%%%%%%%%%%%%%%%%%%%%%%%%%%%%%%%%%%%%%%%%%%%%%%%%%%%%%%%%%%%%%%%%%%%%%%%%%%%%%

\subsection{Qualitative Analysis}\label{sec:quality}
To qualitatively reveal the characteristics of QPIC and the main reasons behind its superior performance over existing methods, we analyze the failure cases of existing methods and QPIC's behavior in the cases.
The top row in Fig.~\ref{fig:fig4_qualitative} shows the failure cases shown in Fig.~\ref{fig:fig1_heatmap}, and the bottom row illustrates the attentions of QPIC on the images.

Figure~\ref{fig:fig4_qualitative_a} and~\ref{fig:fig4_qualitative_b} show the cases where DRG fails to detect the action classes, but QPIC does not. 
As previously discussed, the regions in an image other than a human and object bounding box sometimes contain useful information.
Figure~\ref{fig:fig4_qualitative_a} is a typical example, where the hose held by the boy is likely to be the important contextual information.
Two-stage methods that utilize only the region features, namely the human and object bounding box (and sometimes the union region of the two), cannot fully leverage the contextual information, whereas QPIC successfully places the distinguishing focus on such information and leverages it as shown in the attention map.
Furthermore, the region features are sometimes contaminated by other region features when target bounding boxes are overlapped. 
Figure~\ref{fig:fig4_qualitative_b} shows such an example, where the hand of the blocking man is contained in the bounding box of the catching man.
The typical two-stage methods, which rely on region features, cannot exclude this disturbing information, resulting in incorrect detection.
QPIC, however, can selectively aggregate only the helpful information for each HOI as shown in the attention map, resulting in the correct detection.

Figure~\ref{fig:fig4_qualitative_c} and~\ref{fig:fig4_qualitative_d} illustrate the failure cases of PPDM, whose detection points are drawn in yellow circles. 
As discussed in Sec.~\ref{subsec:sota}, features of heuristic detection points are sometimes dominated by irrelevant information such as the non-target human in Fig.~\ref{fig:fig4_qualitative_c} and another HOI features in Fig.~\ref{fig:fig4_qualitative_d}. 
Consequently, the detection based on those confusing features tends to result in failures.
QPIC alleviates this problem by incorporating the attention mechanism that selectively captures image-wide features as shown in the attention maps, and thus correctly detects these HOIs.

Overall, these qualitative analysis demonstrates the QPIC's capability of acquiring image-wide contextual features, which lead to its superior performance over the existing methods. 

\section{Conclusion}
We have proposed QPIC, a novel detector that can selectively aggregate image-wide contextual information for HOI detection. QPIC leverages an attention mechanism to effectively aggregate features for detecting a wide variety of HOIs. 
This aggregation enriches HOI features, and as a result, simple and intuitive detection heads are realized.
The evaluation on two benchmark datasets showed QPIC's significant superiority over existing methods. 
The extensive analysis showed that the attention mechanism and query-based detection play a crucial role for HOI detection.

{\small
\bibliographystyle{ieee_fullname}
\bibliography{egbib}

\begin{thebibliography}{10}\itemsep=-1pt

\bibitem{bello_iccv2019}
Irwan Bello, Barret Zoph, Ashish Vaswani, Jonathon Shlens, and Quoc~V. Le.
\newblock Attention augmented convolutional networks.
\newblock In {\em ICCV}, October 2019.

\bibitem{carion_eccv2020}
Nicolas Carion, Francisco Massa, Gabriel Synnaeve, Nicolas Usunier, Alexander
  Kirillov, and Sergey Zagoruyko.
\newblock End-to-end object detection with transformers.
\newblock In {\em ECCV}, September 2020.

\bibitem{chao_wacv2018}
Yu-Wei Chao, Yunfan Liu, Michael Liu, Huayi Zeng, and Jia Deng.
\newblock Learning to detect human-object interactions.
\newblock In {\em WACV}, March 2018.

\bibitem{gao_eccv2020}
Chen Gao, Jiarui Xu, Yuliang Zou, and Jia-Bin Huang.
\newblock {DRG}: Dual relation graph for human-object interaction detection.
\newblock In {\em ECCV}, August 2020.

\bibitem{gao_bmvc2018}
Chen Gao, Yuliang Zou, and Jia-Bin Huang.
\newblock i{CAN}: Instance-centric attention network for human-object
  interaction detection.
\newblock In {\em BMVC}, September 2018.

\bibitem{gkioxari_cvpr2018}
Georgia Gkioxari, Ross Girshick, Piotr Doll{\'a}r, and Kaiming He.
\newblock Detecting and recognizing human-object interactions.
\newblock In {\em CVPR}, June 2018.

\bibitem{gupta_arxiv2015}
Saurabh Gupta and Jitendra Malik.
\newblock Visual semantic role labeling.
\newblock May 2015.
\newblock arXiv:1505.04474.

\bibitem{gupta_iccv2019}
Tanmay Gupta, Alexander Schwing, and Derek Hoiem.
\newblock No-frills human-object interaction detection: Factorization, layout
  encodings, and training techniques.
\newblock In {\em ICCV}, October 2019.

\bibitem{he_iccv2017}
Kaiming He, Georgia Gkioxari, P. Doll{\'a}r, and Ross~B. Girshick.
\newblock Mask {R-CNN}.
\newblock In {\em ICCV}, October 2017.

\bibitem{he_cvpr2016}
Kaiming He, Xiangyu Zhang, Shaoqing Ren, and Jian Sun.
\newblock Deep residual learning for image recognition.
\newblock In {\em CVPR}, June 2016.

\bibitem{zhi_eccv2020}
Zhi Hou, Xiaojiang Peng, Yu Qiao, and Dacheng Tao.
\newblock Visual compositional learning for human-object interaction detection.
\newblock In {\em ECCV}, August 2020.

\bibitem{kim_bumsoo_eccv2020}
Bumsoo Kim, Taeho Choi, Jaewoo Kang, and Hyunwoo~J. Kim.
\newblock Union{D}et: Union-level detector towards real-time human-object
  interaction detection.
\newblock In {\em ECCV}, August 2020.

\bibitem{kim_dong_eccv2020}
Dong-Jin Kim, Xiao Sun, Jinsoo Choi, Stephen Lin, and In~So Kweon.
\newblock Detecting human-object interactions with action co-occurrence priors.
\newblock In {\em ECCV}, August 2020.

\bibitem{kuhn_1955}
H.~W. Kuhn and Bryn Yaw.
\newblock The hungarian method for the assignment problem.
\newblock {\em Naval Res. Logist. Quart}, pages 83--97, 1955.

\bibitem{li_cvpr2020}
Yong-Lu Li, Xinpeng Liu, Han Lu, Shiyi Wang, Junqi Liu, Jiefeng Li, and Cewu
  Lu.
\newblock Detailed 2d-3d joint representation for human-object interaction.
\newblock In {\em CVPR}, June 2020.

\bibitem{li_cvpr2019}
Yong-Lu Li, Siyuan Zhou, Xijie Huang, Liang Xu, Ze Ma, Hao-Shu Fang, Yanfeng
  Wang, and Cewu Lu.
\newblock Transferable interactiveness knowledge for human-object interaction
  detection.
\newblock In {\em CVPR}, June 2019.

\bibitem{liao_cvpr2020}
Yue Liao, Si Liu, Fei Wang, Yanjie Chen, Chen Qian, and Jiashi Feng.
\newblock {PPDM}: Parallel point detection and matching for real-time
  human-object interaction detection.
\newblock In {\em CVPR}, June 2020.

\bibitem{lin_iccv2017}
Tsung-Yi Lin, Priya Goyal, Ross Girshick, Kaiming He, and Piotr Doll{\'a}r.
\newblock Focal loss for dense object detection.
\newblock In {\em ICCV}, October 2017.

\bibitem{lin_ijcai2020}
Xue Lin, Qi Zou, and Xixia Xu.
\newblock Action-guided attention mining and relation reasoning network for
  human-object interaction detection.
\newblock In {\em IJCAI}, July 2020.

\bibitem{liu_eccv2020}
Yang Liu, Qingchao Chen, and Andrew Zisserman.
\newblock Amplifying key cues for human-object-interaction detection.
\newblock In {\em ECCV}, August 2020.

\bibitem{loshchiloy_iclr2019}
Ilya Loshchilov and Frank Hutter.
\newblock Decoupled weight decay regularization.
\newblock In {\em ICLR}, May 2019.

\bibitem{parmar_icml2018}
Niki Parmar, Ashish Vaswani, Jakob Uszkoreit, Lukasz Kaiser, Noam Shazeer,
  Alexander Ku, and Dustin Tran.
\newblock Image transformer.
\newblock In {\em ICML}, July 2018.

\bibitem{peszke_nips2019}
Adam Paszke, Sam Gross, Francisco Massa, Adam Lerer, James Bradbury, Gregory
  Chanan, Trevor Killeen, Zeming Lin, Natalia Gimelshein, Luca Antiga, Alban
  Desmaison, Andreas Kopf, Edward Yang, Zachary DeVito, Martin Raison, Alykhan
  Tejani, Sasank Chilamkurthy, Benoit Steiner, Lu Fang, Junjie Bai, and Soumith
  Chintala.
\newblock Py{T}orch: An imperative style, high-performance deep learning
  library.
\newblock In {\em NeurIPS}, December 2019.

\bibitem{qi_eccv2018}
Siyuan Qi, Wenguan Wang, Baoxiong Jia, Jianbing Shen, and Song-Chun Zhu.
\newblock Learning human-object interactions by graph parsing neural networks.
\newblock In {\em ECCV}, September 2018.

\bibitem{ren_nips2015}
Shaoqing Ren, Kaiming He, Ross Girshick, and Jian Sun.
\newblock Faster {R-CNN}: Towards real-time object detection with region
  proposal networks.
\newblock In {\em NIPS}, December 2015.

\bibitem{rezatofighi_cvpr2019}
Hamid Rezatofighi, Nathan Tsoi, JunYoung Gwak, Amir Sadeghian, Ian Reid, and
  Silvio Savarese.
\newblock Generalized intersection over union: A metric and a loss for bounding
  box regression.
\newblock In {\em CVPR}, June 2019.

\bibitem{ulutan_cvpr2020}
Oytun Ulutan, A~S~M Iftekhar, and B.~S. Manjunath.
\newblock {VSGN}et: Spatial attention network for detecting human object
  interactions using graph convolutions.
\newblock In {\em CVPR}, June 2020.

\bibitem{vaswani_nips2017}
Ashish Vaswani, Noam Shazeer, Niki Parmar, Jakob Uszkoreit, Llion Jones,
  Aidan~N Gomez, \L~ukasz Kaiser, and Illia Polosukhin.
\newblock Attention is all you need.
\newblock In {\em NIPS}, December 2017.

\bibitem{wan_iccv2019}
Bo Wan, Desen Zhou, Yongfei Liu, Rongjie Li, and Xuming He.
\newblock Pose-aware multi-level feature network for human object interaction
  detection.
\newblock In {\em ICCV}, October 2019.

\bibitem{hai_eccv2020}
Hai Wang, Wei shi Zheng, and Ling Yingbiao.
\newblock Contextual heterogeneous graph network for human-object interaction
  detection.
\newblock In {\em ECCV}, August 2020.

\bibitem{wang_iccv2019}
Tiancai Wang, Rao~Muhammad Anwer, Muhammad~Haris Khan, Fahad~Shahbaz Khan,
  Yanwei Pang, Ling Shao, and Jorma Laaksonen.
\newblock Deep contextual attention for human-object interaction detection.
\newblock In {\em ICCV}, October 2019.

\bibitem{wang_cvpr2020}
Tiancai Wang, Tong Yang, Martin Danelljan, Fahad~Shahbaz Khan, Xiangyu Zhang,
  and Jian Sun.
\newblock Learning human-object interaction detection using interaction points.
\newblock In {\em CVPR}, June 2020.

\bibitem{xu_tmm2020}
Bingjie Xu, Junnan Li, Yongkang Wong, Qi Zhao, and Mohan~S. Kankanhalli.
\newblock Interact as you intend: Intention-driven human-object interaction
  detection.
\newblock {\em TMM}, 22(6):1423--1432, June 2020.

\bibitem{yang_ijcai2020}
Dongming Yang and Yuexian Zou.
\newblock A graph-based interactive reasoning for human-object interaction
  detection.
\newblock In {\em IJCAI}, July 2020.

\bibitem{zhong_eccv2020}
Xubin Zhong, Changxing Ding, Xian Qu, and Dacheng Tao.
\newblock Polysemy deciphering network for robust human-object interaction
  detection.
\newblock In {\em ECCV}, August 2020.

\bibitem{zhou_iccv2019}
Penghao Zhou and Mingmin Chi.
\newblock Relation parsing neural network for human-object interaction
  detection.
\newblock In {\em ICCV}, October 2019.

\bibitem{zhou_cvpr2020}
Tianfei Zhou, Wenguan Wang, Siyuan Qi, Haibin Ling, and Jianbing Shen.
\newblock Cascaded human-object interaction recognition.
\newblock In {\em CVPR}, June 2020.

\bibitem{zhou_center_arxiv2019}
Xingyi Zhou, Dequan Wang, and Philipp Kr{\"a}henb{\"u}hl.
\newblock Objects as points, April 2019.
\newblock arXiv:1904.07850.

\end{thebibliography}
}

\renewcommand{\thesection}{\Alph{section}}
\setcounter{section}{0}

\clearpage

\section{Supplementary V-COCO Settings}
As mentioned in the main manuscript, the images of the V-COCO dataset are split into three sets: a training set, validation set, and testing set. Following previous works, the training and validation sets are combined to train QPIC. 

For calculating the mAP, 5 action classes out of the 29 classes are excluded from the evaluation following~\cite{gkioxari_cvpr2018}. This is because four of the excluded action classes (``run", ``smile", ``stand", and ``walk") are the action without an object, and one of them (``point") has an insufficient number of samples.

\section{Supplementary Implementation Note}
As usual training, we use data augmentation to alleviate over-fitting. We use random horizontal flipping augmentation, scale augmentation, random crop augmentation, which are used in DETR's training~\cite{carion_eccv2020}, and color augmentation, which is used in PPDM's training~\cite{liao_cvpr2020}.

Since each layer of a transformer decoder output its own set of embeddings $\bm{D} = \{\bm{d}_{i} | \bm{d}_{i} \in \mathbb{R}^{D_{c}}\}_{i=1}^{N_{q}}$, the loss calculation described in Sec.~{3.2} of the main manuscript can be conducted for each layer. Following the DETR's training~\cite{carion_eccv2020}, these auxiliary losses are calculated to optimize QPIC. To calculate the losses, FFNs are added on top of each decoder layer's output. Note that the parameters of the FNNs are shared among all the decoder layers, 

In the evaluation time, the second highest scoring class and confidence of the object-class prediction $\bm{\hat{c}}_{i}$ are used to generate the detection result if $\bm{\hat{c}}_{i}$ has the highest score in ``no pair" class. This is the technique used in~\cite{carion_eccv2020} to optimize the mAPs.

\section{Additional List of Comparison}
\begin{table}[t]
    \caption{Comparison against state-of-the-art methods on HICO-DET. The top, middle, and bottom blocks show the mAPs of the two-stage, single-stage, and our methods, respectively.}
    \label{table:comp_hico}
    \centering
    \small
    \setlength{\tabcolsep}{2pt}
    \begin{tabular}{@{}lcccccc@{}}
        \toprule
        & \multicolumn{3}{c}{Default} & \multicolumn{3}{c}{Known object} \\
        \cmidrule(lr){2-4}\cmidrule(lr){5-7}
        Method & full & rare & non-rare & full & rare & non-rare \\
        \midrule
        PMFNet~\cite{wan_iccv2019} & 17.46 & 15.65 & 18.00 & 20.34 & 17.47 & 21.20 \\
        Wang \etal~\cite{hai_eccv2020} & 17.57 & 16.85 & 17.78 & 21.00 & 20.74 & 21.08 \\
        In-GraphNet~\cite{yang_ijcai2020} & 17.72 & 12.93 & 19.31 & -- & -- & -- \\
        VSGNet~\cite{ulutan_cvpr2020} & 19.80 & 16.05 & 20.91 & -- & -- & -- \\
        FCMNet~\cite{liu_eccv2020} & 20.41 & 17.34 & 21.56 & 22.04 & 18.97 & 23.13 \\
        ACP~\cite{kim_dong_eccv2020} & 20.59 & 15.92 & 21.98 & -- & -- & -- \\
        PD-Net~\cite{zhong_eccv2020} & 20.81 & 15.90 & 22.28 & 24.78 & 18.88 & 26.54 \\
        DJ-RM~\cite{li_cvpr2020} & 21.34 & 18.53 & 22.18 & 23.69 & 20.64 & 24.60 \\
        VCL~\cite{zhi_eccv2020} & 23.63 & 17.21 & 25.55 & 25.98 & 19.12 & 28.03 \\
        DRG~\cite{gao_eccv2020} & 24.53 & 19.47 & 26.04 & 27.98 & 23.11 & 29.43 \\
        \midrule
        UnionDet~\cite{kim_bumsoo_eccv2020} & 17.58 & 11.72 & 19.33 & 19.76 & 14.68 & 21.27 \\
        Wang \etal~\cite{wang_cvpr2020} & 19.56 & 12.79 & 21.58 & 22.05 & 15.77 & 23.92 \\
        PPDM~\cite{liao_cvpr2020} & 21.73 & 13.78 & 24.10 & 24.58 & 16.65 & 26.84 \\
        \midrule
        Ours (ResNet-50) & 29.07 & 21.85 & 31.23 & 31.68 & 24.14 & 33.93 \\
        Ours (ResNet-101) & \textbf{29.90} & \textbf{23.92} & \textbf{31.69} & \textbf{32.38} & \textbf{26.06} & \textbf{34.27} \\
        \bottomrule
    \end{tabular}
\end{table}

\begin{table}[t]
    \caption{Comparison against state-of-the-art methods on V-COCO. The split of the blocks are the same as Table~\ref{table:comp_hico}.}
    \label{table:comp_vcoco}
    \centering
    \small
    \begin{tabular}{@{}lcc@{}}
        \toprule
        Method & Scenario 1 & Scenario 2 \\
        \midrule
        VCL~\cite{zhi_eccv2020} & 48.3 & -- \\
        In-GraphNet~\cite{yang_ijcai2020} & 48.9 & -- \\
        DRG~\cite{gao_eccv2020} & 51.0 & -- \\
        VSGNet~\cite{ulutan_cvpr2020} & 51.8 & 57.0 \\
        PMFNet~\cite{wan_iccv2019} & 52.0 & -- \\
        PD-Net~\cite{zhong_eccv2020} & 52.6 & -- \\
        Wang \etal~\cite{hai_eccv2020} & 52.7 & -- \\
        ACP~\cite{kim_dong_eccv2020} & 53.0 & -- \\
        FCMNet~\cite{liu_eccv2020} & 53.1 & -- \\
        \midrule
        UnionDet~\cite{kim_bumsoo_eccv2020} & 47.5 & 56.2 \\
        Wang \etal~\cite{wang_cvpr2020} & 51.0 & -- \\
        \midrule
        Ours (ResNet-50) & \textbf{58.8} & \textbf{61.0} \\
        Ours (ResNet-101) & 58.3 & 60.7\\
        \bottomrule
    \end{tabular}
    \vspace{-2.0ex}
\end{table}

\begin{figure*}[t]
\centering
\subfloat[]{%
    \label{fig:qualitative_a}
    \includegraphics[clip,keepaspectratio,width=0.23\textwidth]{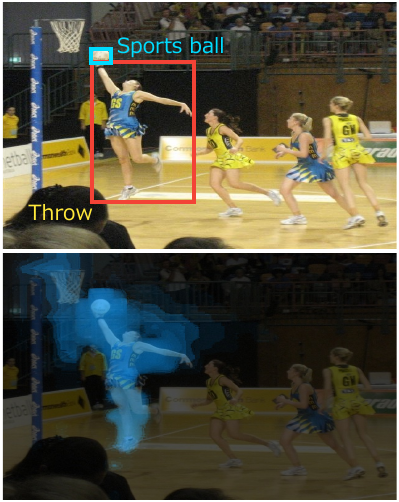}%
}
\subfloat[]{%
    \label{fig:qualitative_b}
    \includegraphics[clip,keepaspectratio,width=0.23\textwidth]{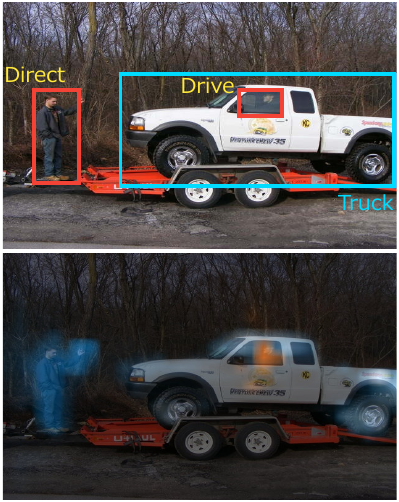}%
}
\subfloat[]{%
    \label{fig:qualitative_c}
    \includegraphics[clip,keepaspectratio,width=0.23\textwidth]{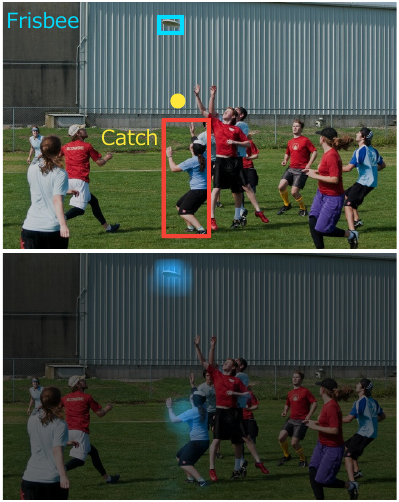}%
}
\subfloat[]{%
    \label{fig:qualitative_d}
    \includegraphics[clip,keepaspectratio,width=0.23\textwidth]{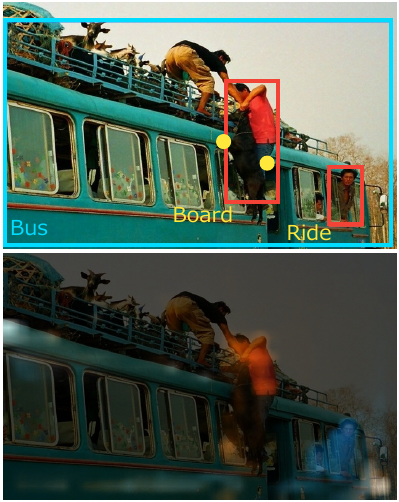}%
}
\caption{Typical failure cases of conventional detectors (top row) and attentions of QPIC (bottom row). The ground-truth human bounding boxes, object bounding boxes, object classes, and action classes are drawn with red boxes, blue boxes, blue characters, and yellow characters, respectively. In (b) and (d), the attentions corresponding to different HOI instances are drawn with blue and orange, and the areas where two attentions overlap are drawn with white.}
\label{fig:qualitative}
\vspace{-2.0ex}
\end{figure*}

Table~\ref{table:comp_hico} and Table~\ref{table:comp_vcoco} show the additional list of the comparison against state-of-the-art on HICO-DET~\cite{chao_wacv2018} and V-COCO~\cite{gupta_arxiv2015}, respectively. Six methods (PMFNet~\cite{wan_iccv2019}, Wang \etal~\cite{hai_eccv2020}, In-GraphNet~\cite{yang_ijcai2020}, VSGNet~\cite{ulutan_cvpr2020}, PD-Net~\cite{zhong_eccv2020}, and DJ-RM~\cite{li_cvpr2020}) are additionally compared in these tables. As stated in Sec.~{4.3} of the main manuscript, our QPIC significantly outperforms conventional two- and single-stage methods on both datasets.

\section{Computational Efficiency Comparison}
\begin{table}[t]
    \caption{Comparison of the efficiency.}
    \label{table:speed}
    \centering
    \small
    \setlength{\tabcolsep}{2pt}
    \begin{tabular}{@{}lcc@{}}
        \toprule
        Method & HICO-DET (mAP) & Inference time (ms) \\
        \midrule
        PPDM~\cite{liao_cvpr2020} & 21.73 & 64 \\
        Ours (ResNet-50) & 29.07 & \textbf{46} \\
        Ours (ResNet-101) & \textbf{29.90} & 63 \\
        \bottomrule
    \end{tabular}
\end{table}

To analyze the model efficiency of our QPIC, we compare the inference times of QPIC and PPDM~\cite{liao_cvpr2020}, which is one of the highest speed models. We used the publicly available source code of PPDM\footnote{https://github.com/YueLiao/PPDM}, and tested each model on a single Tesla V100 GPU with CUDA ver.~{10.1} and PyTorch ver.~{1.5}~\cite{peszke_nips2019}. Table~\ref{table:speed} shows the comparison result. As the table shows, the inference time of QPIC with the ResNet-50 backbone is smaller by 18 ms than that of PPDM. In particular, PPDM takes 17 ms to organize outputs from the network, while QPIC takes only 5.4 ms to do that. These results indicate that QPIC is more efficient than conventional methods mainly because the simple detection heads of QPIC realize the simple inference procedures.

\section{Additional Qualitative Analysis}
Figure~\ref{fig:qualitative} shows the additional failure cases of conventional methods. Figure~\ref{fig:qualitative_a} and~\ref{fig:qualitative_b} show the failure cases of DRG~\cite{gao_eccv2020}, and Fig.~\ref{fig:qualitative_c} and~\ref{fig:qualitative_d} show those of PPDM~\cite{liao_cvpr2020}, where QPIC successfully detects the human-object interactions (HOIs).
As discussed in the main manuscript, the regions in an image other than a human and object bounding box sometimes contain useful information. Fig.~\ref{fig:qualitative_a} is a typical example case, where the basketball goal is likely to be the important contextual information. The attention of QPIC shows that it aggregates features from the region of the basketball goal, resulting in the correct detection.
Figure~\ref{fig:qualitative_b} shows an example case where multiple HOI instances are overlapped. As shown in the figure, the bounding box of the track includes that of the driving human, which may induce contaminated features. The performance is degraded by this contamination. Unlike DRG, QPIC selectively aggregates features for each HOI using the attention mechanism as shown in the attention map, and successfully detects the HOIs.
In Fig.~\ref{fig:qualitative_c} and~\ref{fig:qualitative_d}, the features of the detection points, which are the locations to predict HOIs in PPDM and drawn in the yellow circles in the figures, are likely to be dominated by irrelevant information because the points are on the background or irrelevant human. As is the case with DRG, PPDM cannot predict HOIs with these contaminated features, while QPIC can do it with the selectively aggregated features.

\end{document}